\documentclass{article}
\usepackage{spconf,amsmath,graphicx}

\title{TOWARDS DEEPER GENERATIVE ARCHITECTURES FOR GANS USING DENSE CONNECTIONS}
\name{Samarth Tripathi, Renbo Tu}
\address{Columbia University \\ Dept of Computer Science \\ New York, NY 10027}
\begin{document}
%
\maketitle
\begin{abstract}
  In this paper, we present the result of adopting skip connections and dense layers, previously used in image classification tasks, to the Fisher GAN implementation. We have experimented with different numbers of layers and inserting these connections in different sections of the network. Our findings suggests that generative networks implemented with skip connections produce better images than the baseline, and the number of connections added has a varied effect on the result. 
\end{abstract}
\section{Introduction}
\label{sec:intro}

Recent developments in Generative Adversarial Networks (GANs) have allowed for production of high-quality, quasi-natural images. In this research, we seek to develop another variation of GAN to generate such semi-realistic images. Considering the significant progress that took place in the image classification field, specifically with DenseNet achieving state-of-the-art performance, we seek to leverage the strength of the dense networks to improve GAN performance. Our model adopts the DenseNet and ResNet architecture variations as the generator, which is modified to include the skip connections found in these classification networks. We expect these connections to enable the generator to learn more complex features of the image, as skip connections/dense connections have proved crucial for extracting complex features in classification tasks. On the other hand, since converging has proved a difficulty for GANs, we used Fisher GAN as our base model to improve convergence performances. We have implemented the models for two datasets: CelebA and Cifar10.

In this paper, we make the following contributions:
\begin{itemize}
\item 
We use the newly constructed model for generating higher quality images based on common evaluation metrics such as inception score.

\item 
We analyze the effect of skip connections in generating images through comparing models with different dense architectures.

\item
We investigate in the optimal training method for these GANs with dense connections.

\end{itemize}

\section{Related Works}
\label{sec:RW}

\begin{enumerate}
    \item 
    Fisher GAN: 
    As GANs have proven to be unstable during training, attaining convergence for the generator has been one of the primary difficulties for researchers. The newly developed Fisher GAN (Mroueh \& Sercu, 2017 \cite{DBLP:journals/corr/MrouehS17}) defines a critic with a data dependent constraint on its second order moments. The new algorithm based on the Augmented Lagrangian, incorporated in a DCGAN (Radford et. al., 2016 \cite{DBLP:journals/corr/RadfordMC15}) network, achieved robust performance in terms of the semi-supervised learning metric and generated good samples.

    \item
    DenseNet:
    DenseNet (Huang et.al., 2016 \cite{DBLP:journals/corr/HuangLW16}) is a deep convolutional network with skip connections between each layer to every other layer within the same block in a feed-forward fashion. Traditional ResNet (He et.al., 2015 \cite{DBLP:journals/corr/HeZRS15}) has showed that residual layers with skip connections can learn high-level features more accurately and gain accuracy with increased depth. The DenseNet implementation successfully addresses the vanishing gradient problem, strengthens feature learning, and reduces the number of parameters through bottleneck layers. DenseNet obtained state-of-the-art performance on classification tasks with multiple popular datasets. 
    
    \item 
    Inception Score:
    Salimans et al. (2016 \cite{DBLP:journals/corr/SalimansGZCRC16}) proposes a widely used metric, namely "Inception Score" for evaluating GAN-generated images. The procedure starts with feeding images to an pretrained inception model to obtain conditional model distribution $p(y|x)$. A critic then evaluates this distribution to check if the images contain meaningful objects, which is represented by low entropy. Another objective of the critic judges whether the generated images are varied through evaluating marginal $\int p(y|x = G(z))$, which should have high entropy. This metric is used in conjunction with human judgment (visual inspection). 
    
\end{enumerate}

\section{Experimental Setup}
\label{sec:ES}

\begin{figure}[h]
\caption{Diagram depicting connections within our model Architectures }
\includegraphics[width=1\linewidth]{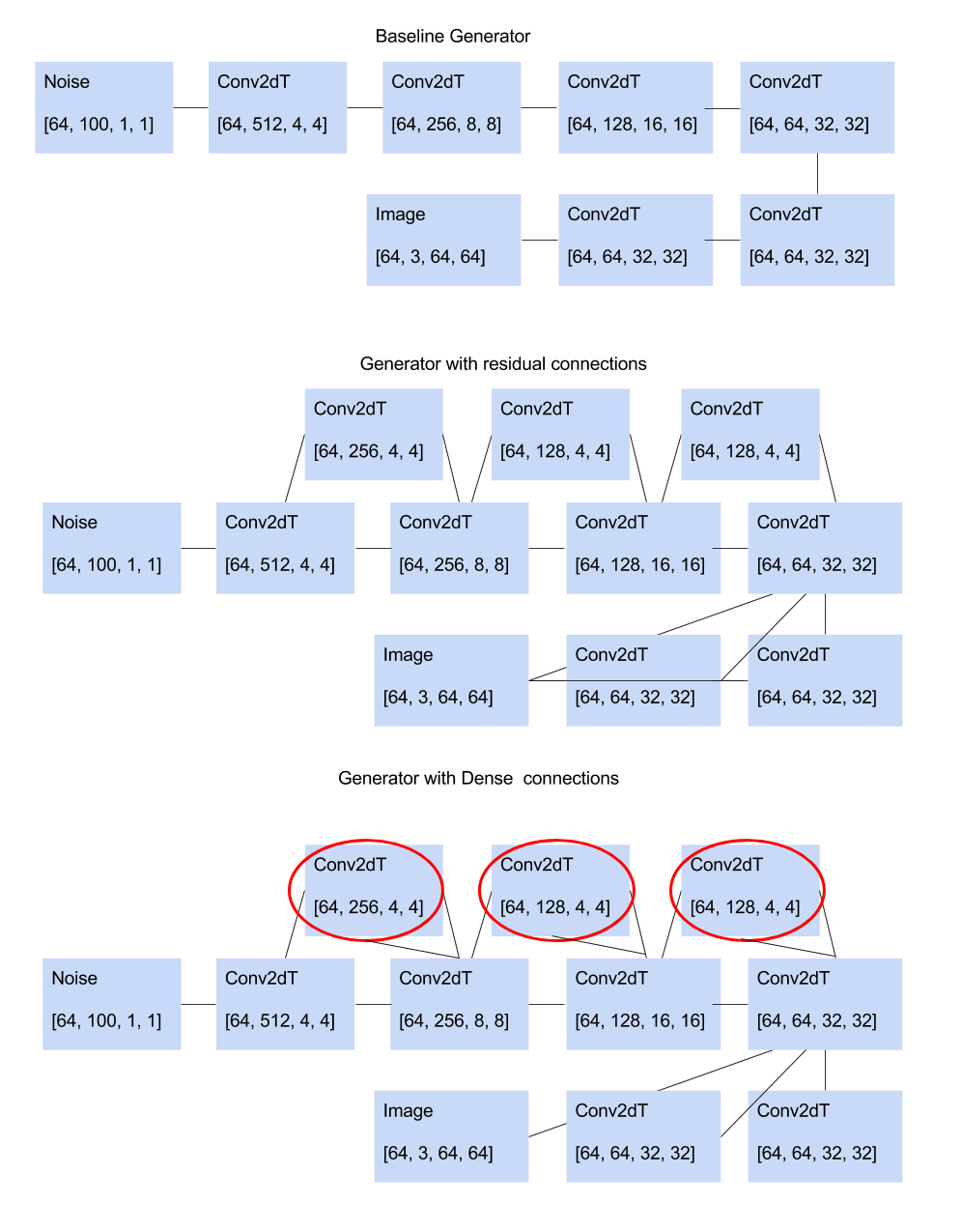}
\end{figure}

We first discuss the DCGAN structure we try to improve upon. The architecture is presented in figure 1, which consists of log2 image size layers (6 layers and 5 layer for image size of 64 and 32 pixels) respectively of Convolution layers of 4*4 size with a stride of 2 and padding of 1 for the Discriminator (and Convolution-Transpose for the Generator). These "essential" layers perform dimensionality reduction (and expansion) by a factor of 2 and are indispensable for the models. Apart from these the model also contains “extra layers”, which perform basic Convolution (and Convolution-Transpose) layers of 3*3 size with a stride of 1 and padding of 1. FisherGAN uses 2 extra layers each for Discriminator and Generator. The layers also use Batchnormalization as a precursor to each Convolutional layer, followed by ReLu for Generators and LeakyReLu for Discriminator. There are further many hyperparameters and constraints mentioned in the FisherGAN paper that we replicate for both CelebA and Cifar10.

For our research we only experiment adding residual and dense connections to the Generator model, and don’t interfere with discriminator. The reason for this approach is two fold. Firstly, it becomes very difficult to find the correct hyperparameters and constraints to effectively train these models, which becomes increasing difficult as we increase the number of layers in a generic manner. Secondly, we do not notice improvements and only find the performance to deteriorate compared to the baseline FisherGan. To improve upon the DCGAN structure, we first add Residual connections between layers in Generative models, while keeping the discriminator constant. This allows us to train the GAN using the same hyperparameters and constraints as DCGAN while allowing deeper and more powerful generators. We add 5 Transpose-Convolution layers like extra layers, but with residual connections between DCGAN layers in the form of Figure 1. These residual layers keep the dimensionality constant which allows us to concatenate the feature maps before sending them to the next layer.

Next we try to add more layers and make the generators deeper in a Densely Connected manner. We achieve this by adding more Transpose-Convolution which takes as input all the previous layer’s output (within the same block) and outputs the concatenated feature maps of all inputs and its outputs to the next layer (keeping the same output dimensionality). As we go deeper, adding more layers results in an exponential increase in feature maps being passed on the next layer, which increases computation and decreases performance. We avoid this by decreasing the number of activation maps from each subsequent dense layer by a factor of two after the first dense layer. This limits both the number of layer to which we can extend and also decreases the number of activation maps that get forwarded to the next layer.

As mentioned our proposed technique decreases the number of activation maps from each subsequent dense layer by a factor of two after the first dense layer. Fisher GAN architecture with 64 being the batch size-
\begin{enumerate}
    \item Input Size([64, 100, 1, 1])
    \item Essential layer Size([64, 256, 4, 4])
    \item Essential layer Size([64, 128, 8, 8])
    \item Essential layer Size([64, 64, 16, 16])
    \item Extra layer Size([64, 64, 16, 16])
    \item Extra layer Size([64, 64, 16, 16])
    \item Essential layer Size([64, 3, 32, 32])
    \item Extra layer Size([64, 3, 32, 32])
\end{enumerate}
Our architecture with dense connections with halved feature maps consecutively.
\begin{enumerate}
    \item Input Size([64, 100, 1, 1])
    \item Essential layer Size([64, 256, 4, 4])
    \item Dense layer1 Size([64, 256, 4, 4])
    \item Dense layer2 Size([64, 128, 4, 4])
    \item Dense layer3 Size([64, 64, 4, 4])
    \item Merged layer Size([64, 704, 4, 4])
    \item Essential layer Size([64, 128, 8, 8])
    \item Dense layer1 Size([64, 128, 8, 8])
    \item Dense layer2 Size([64, 64, 8, 8])
    \item Dense layer3 Size([64, 32, 8, 8])
    \item Merged layer Size([64, 352, 8, 8])
    \item Essential layer Size([64, 64, 16, 16])
    \item Dense layer1 Size([64, 64, 16, 16])
    \item Dense layer2 Size([64, 32, 16, 16])
    \item Dense layer3 Size([64, 16, 16, 16])
    \item Merged layer Size([64, 176, 16, 16])
    \item Essential layer Size([64, 3, 32, 32])
    \item Extra layer Size([64, 3, 32, 32])
\end{enumerate}

We use this architecture for Cifar-10 dataset where images have a size of 32$*$32 pixels. We call this architecture our Dense12-GAN as it has 9 total Dense layers and 3 Merged layers, all of which are evenly distributed between 4 essential layer blocks. For Celeb A which has 64$8$64 sized images we add similarly get Dense16-GAN with 3 dense and 1 merged layers between 5 essential layers.

Our architectures include the baseline Generator(1), Generator with 1 residual connections(Dense4), Generator with 2 Dense Connections(Dense8), Generator with 3 Dense Connections(Dense12), and Generator with 3 Dense connections with decreasing activation maps for Cifar10(Dense16r). This 'Dense16r' architecture involves adding 1D ConvolutionTranspose layer after our Merged layer and before our essential layers to reduce parameters and add deeper convolutions. For example 'Merged layer Size([64, 704, 4, 4])' in line 6, is followed by 1D ConvolutionTranspose layers to change to 'Reduced layer Size([64, 256, 4, 4])'.

As mentioned before naively adding more dense layers ends up deteriorating the performance. We also tried mirroring the Discriminator with dense architectures like the generator network, but we could not effectively train it.

\section{Results and Discussions}
\label{sec:results}

\begin{figure}
\caption{Inception Score Chart}
\includegraphics[width=1\linewidth]{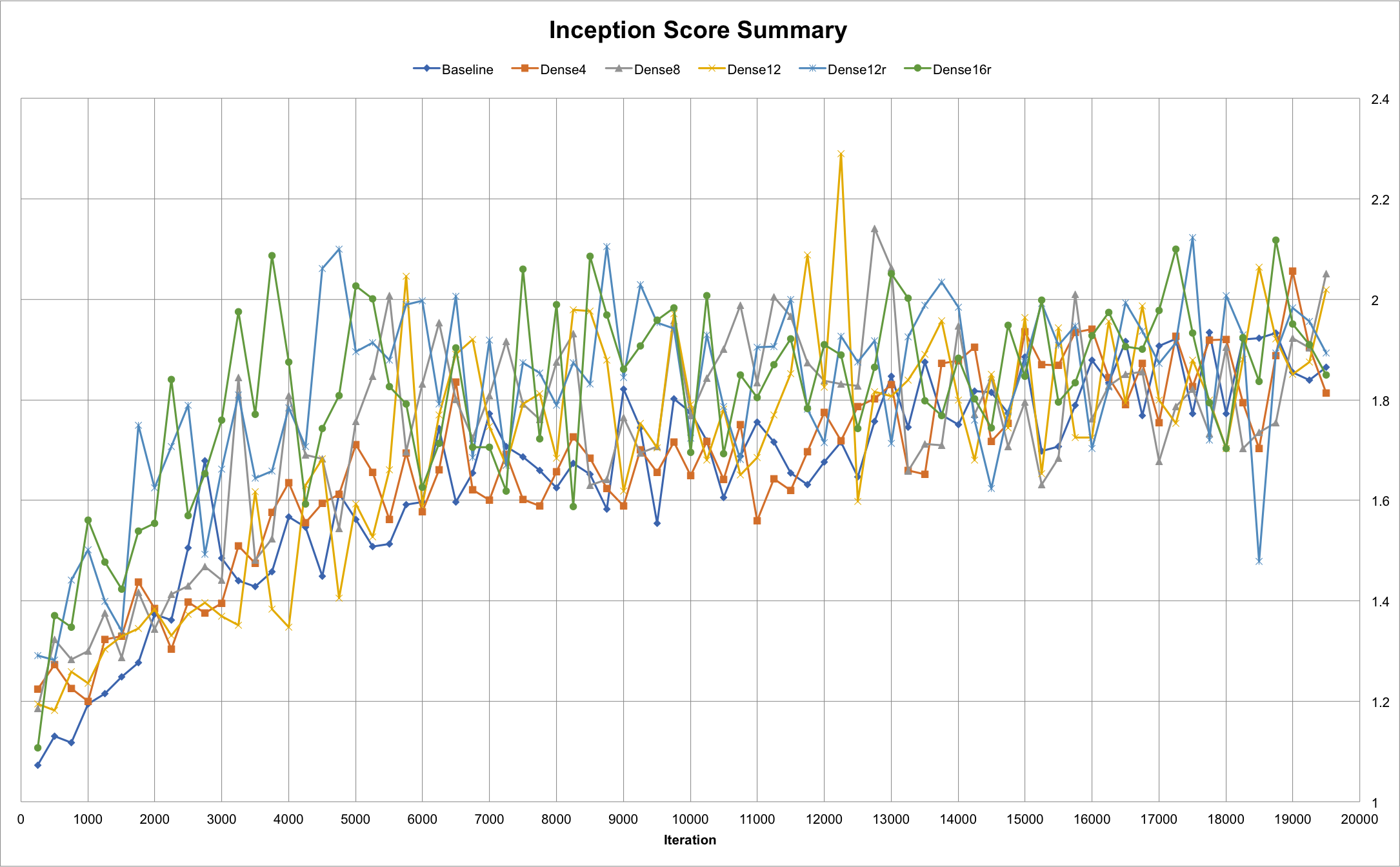}
\end{figure}

\begin{figure*}[h]
\caption{Visual inspection of Celeb A generated Images}
\includegraphics[width=\textwidth]{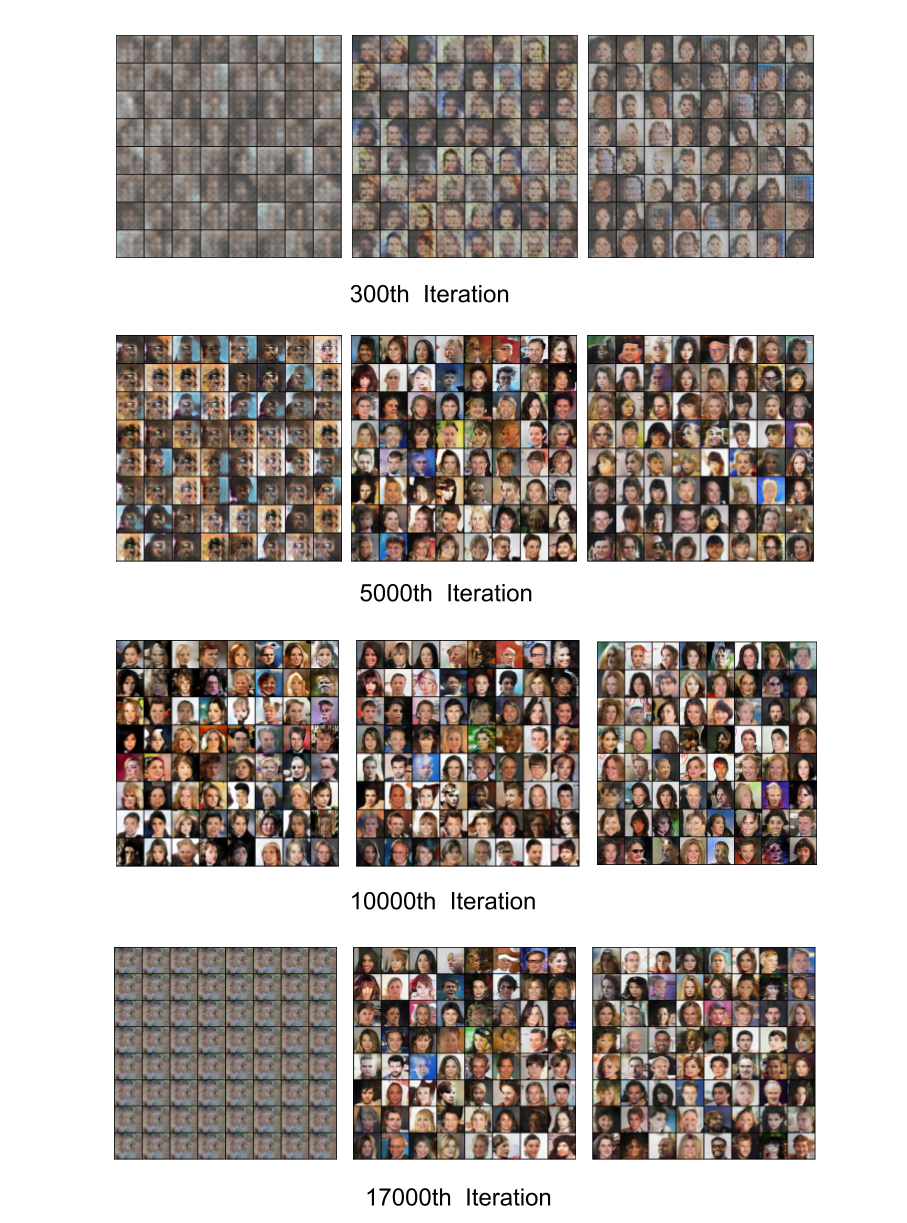}
\end{figure*}

\begin{enumerate}

\item Inception Score Metrics

As previously discussed, inception score has been crucial for result analysis in this research, as it offers a quantitative assessment of the images, both in their diversity and their realistic qualities. The subtle change in the samples, which is not apparent from visual inspection, can be identified when inception score is plotted. 

Figure 2 shows Inception score for Cifar10. The results show substantial improvement over baseline FisherGAN by adding either 1,2,3 Dense layers (Dense4, Dense8, Dense12) between essential layers. We also try to reduce feature maps by adding 1D ConvolutionTranspose (Dense8r, Dense12r) after merged layers to reduce the feature maps back to the original FisherGAN number which keep the quality improvement of DenseGAN with lower computation. Also all the deeper (8,12,12r,16r) Dense models get better scores in general, and achieve better and faster convergence.

\item Visual inspection

We can visually inspect the quality of generated images as well, as shown in Figure 3. with baseline figures on the left, and Dense16 in the middle and Dense20r on the right. The Dense Generator model converge faster, yield more realistic images, and are more robust.

\end{enumerate}

\end{document}